\documentclass[conference,a4paper]{APSIPA2025}

\makeatletter
\def\protected@file@percent{}
\makeatother

\usepackage[T1]{fontenc}
\usepackage[utf8]{inputenc}
\usepackage{url}
\usepackage{fancyhdr}
\usepackage{amsmath, amsfonts, amssymb}
\usepackage{graphicx}
\usepackage{multirow}
\usepackage{threeparttable}
\usepackage{booktabs}
\usepackage{makecell}
\usepackage[backend=biber,style=ieee,]{biblatex}
\usepackage{geometry}
\fancypagestyle{firststyle}{
  \fancyhf{}
  \fancyhead[C]{2025 Asia Pacific Signal and Information Processing Association Annual Summit and Conference (APSIPA ASC)}
}

\begin{document}

\title{Explainable Disentanglement on Discrete Speech Representations for Noise-Robust ASR}

\author{
\authorblockN{
Shreyas Gopal,
Ashutosh Anshul,
Haoyang Li,
Yue Heng Yeo,
Hexin Liu,
Eng Siong Chng
}

\authorblockA{
College of Computing and Data Science, Nanyang Technological University, Singapore \\
E-mail: shreyas011@e.ntu.edu.sg}
}

\maketitle
\pagestyle{empty}
\thispagestyle{firststyle}

\begin{abstract}
Discrete audio representations are gaining traction in speech modeling due to their interpretability and compatibility with large language models, but are not always optimized for noisy or real-world environments. Building on existing works that quantize Whisper embeddings for speech-to-unit modeling, we propose disentangling semantic speech content from background noise in the latent space. Our end-to-end model separates clean speech in the form of codebook tokens, while extracting interpretable noise vectors as quantization residue which are supervised via a lightweight classifier. We show that our approach improves alignment between clean/noisy speech and text, producing speech tokens that display a high degree of noise-invariance, and improves ASR performance. Keeping Whisper frozen, we show an 82\% reduction in error rate compared to Whisper, and 35\% improvement over baseline methods on the VBDemand test set. Further analyses show that the learned token space generalizes well to both seen and unseen acoustic conditions.
\end{abstract}

\vspace{-12 pt}
\section{Introduction}

Recent advances in speech modeling have increasingly adopted discrete audio representations for tasks such as automatic speech recognition (ASR), text-to-speech (TTS), and speech-language modeling. Existing works on neural audio codecs, such as DAC \cite{kumar2023dac}, EnCodec \cite{defossez2022encodec}, and SoundStream \cite{zeghidour2021soundstream}, focus on compressing audio into discrete tokens for high-quality reconstruction and low-bitrate transmission. Works like \cite{dhawan2024codecasr, puvvada2024discrete} demonstrate that with task-supervised training, these discrete representations can also yield competitive ASR results. Similarly, methods like \cite{hono2023integrating, du2024cosyvoice, kyutai2024moshi} have shown that vector-quantized (VQ) tokens can capture compressed and interpretable speech units for TTS and speech generation. This trend is also motivated by the natural compatibility between discrete audio and tokenized sequences used in large language models (LLMs). For example, HuBERT-based speech representations \cite{hsu2021hubert} combined with LLMs can achieve strong ASR performance, as shown in \cite{hono2023integrating, xu2024comparing_disc_cont}. Unfortunately, a key issue with these VQ-based approaches is the information loss introduced by quantization, which can hinder downstream performance, particularly under noisy or adverse conditions.

Most prior works focus on representing clean speech, which does not reflect real-world noisy scenarios, highlighting a gap in noise-robust discrete representations. Although OpenAI's Whisper \cite{whisper} exhibits strong performance under clean or moderately noisy conditions, \cite{gong2023whispernoise, mamba_in_speech} show that its continuous latent representations still encode background noise, which may affect downstream tasks. This issue is echoed in Speechless \cite{dao2025speechless}, which highlights a performance gap between clean and noisy inputs when using a similar VQ-module on Whisper embeddings. These observations suggest that quantized representations align well with clean inputs but fail to generalize well under noisy or real world settings, leading to semantic degradation. We posit that ASR supervision and semantic alignment alone are insufficient to address this issue. Instead, jointly supervising both semantic and noise representations provides a more effective path to robust speech modeling.

Aligned with this hypothesis, we introduce a vector quantization (VQ) module that separates clean speech features from noise in Whisper’s latent space. We freeze the Whisper encoder and decoder and train lightweight modules that align with clean targets while guiding the quantization residue to capture noise. Our key idea is to frame disentanglement as a vector difference: the quantization residue is treated as an explicit and interpretable representation of background noise.

\subsection{Related Works}

Methods such as \cite{wang2022wav2vec, ng2022i2cr} aim to directly improve speech encoder robustness to noise using contrastive losses. Several other approaches attempt to disentangle semantics from noise. Omran et al. \cite{omran2023disentangling} and Bie et al. \cite{bie2025learningsourcedisentanglement} allocate partitions or separate codebooks to semantic and non-semantic components using architectural or label-based supervision. De’hubert \cite{dehubert} applies cross-correlation and contrastive losses on HuBERT embeddings to align semantically similar sentences injected with different noise conditions.

In contrast to approaches that rely on partitioning or multiple codebooks, we use a single codebook to capture speech semantics. Although traditional RVQ methods quantize the residue at multiple stages, our setup uses a single-stage quantizer and interprets the unquantized residue as an explicit estimate of background noise. A lightweight classifier supervises this residue without mapping it to discrete tokens. While De’hubert relies on fine-tuned HuBERT embeddings and shows gains in ASR performance, our system benefits from Whisper’s ASR pretraining, providing a strong semantic prior.

\subsection{Our Contributions}

In summary:
\begin{itemize}
    \item{We propose an end-to-end model that disentangles speech from noise in the latent space of Whisper \cite{whisper} using a vector quantizer that captures semantic features and treats quantization error as noise.}
    \item{We define a compound loss that guides both clean alignment and noise extraction in an explainable manner.}
    \item{We show that the quantization residue is interpretable and can be explicitly classified, while also demonstrating robustness to unseen noise.}
    \item{Despite information loss due to imperfect disentanglement, our method achieves competitive ASR performance, preserving useful semantic information.}
\end{itemize}

\section{Methodology}

\begin{figure}[t]
    \centering
    \includegraphics[width=0.80\linewidth]{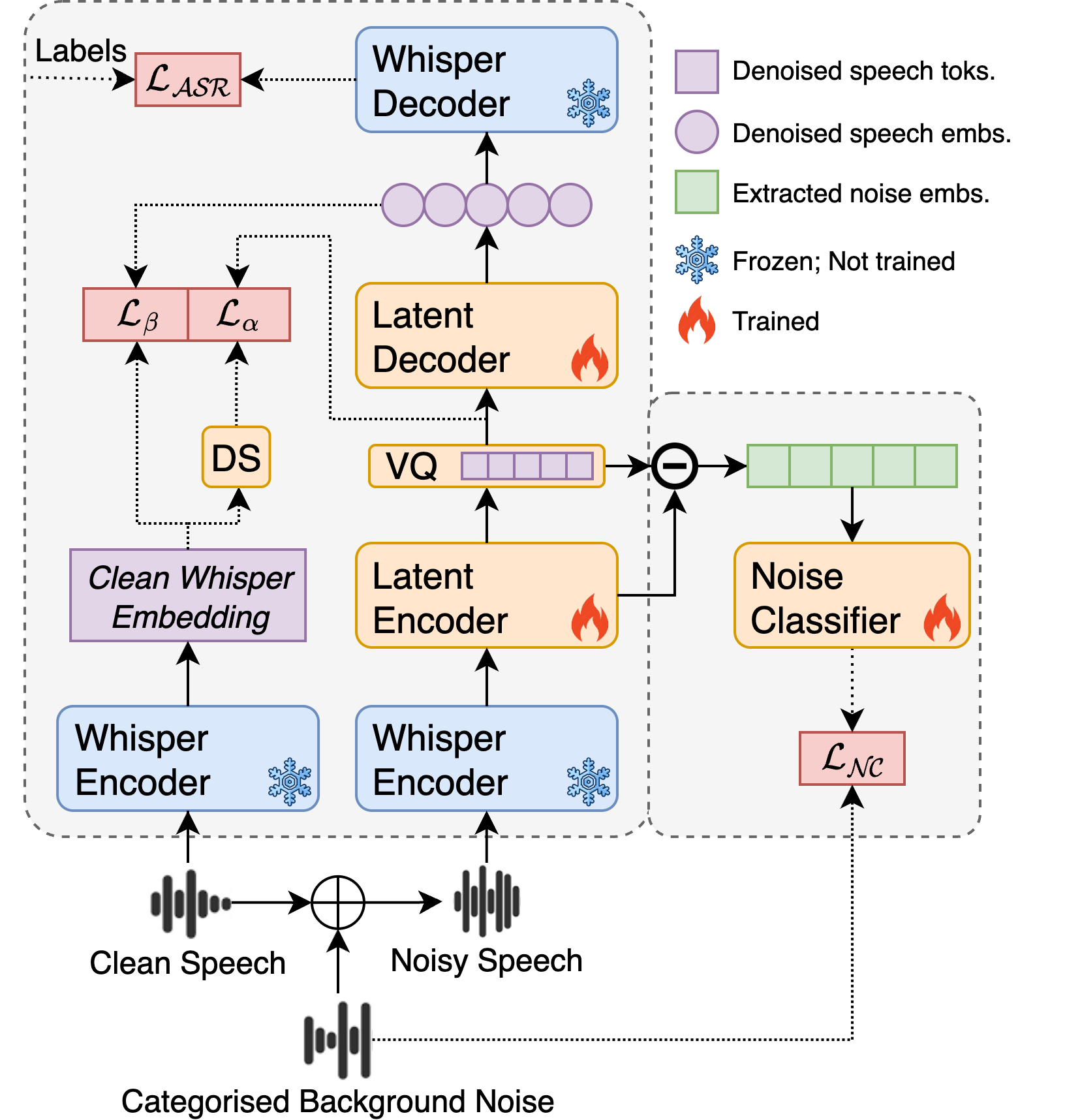}
    \caption{Proposed model architecture. The dotted rectangles highlight two main components: (1) an extended speech encoder (whisper encoder and latent encoder) and latent decoder that improves alignment between clean speech tokens and text, and (2) a noise disentanglement module that guides the quantization residue to model background noise.}
    \label{complete_arch}
\end{figure}

We build on the work of Speechless \cite{dao2025speechless}, a speech-to-unit model using residual vector quantization (RVQ). Our model uses the pretrained \textit{whisper-medium} encoder to provide latent speech embeddings, and the Whisper encoder and decoder remain frozen during training. Fig.~\ref{complete_arch} shows the overall architecture, and in this section we explain the underlying modules.

\subsection{Extended Encoder}

We use paired clean and noisy monophonic signals, $X' \in \mathbb{R}^{1 \times T}$ and $X \in \mathbb{R}^{1 \times T}$, containing the same semantic content. The clean signal is only used for reference in loss computation, while only the noisy signal is passed through the model.

\subsubsection{Whisper Encoder}

The frozen Whisper encoder outputs $W(X'), W(X) \in \mathbb{R}^{T' \times D}$, where $T' = 1500$ denotes the sequence length and $D = 1024$ the embedding dimension.

\subsubsection{Latent Encoder}

\begin{figure}[t]
    \centering
    \includegraphics[width=0.80\linewidth]{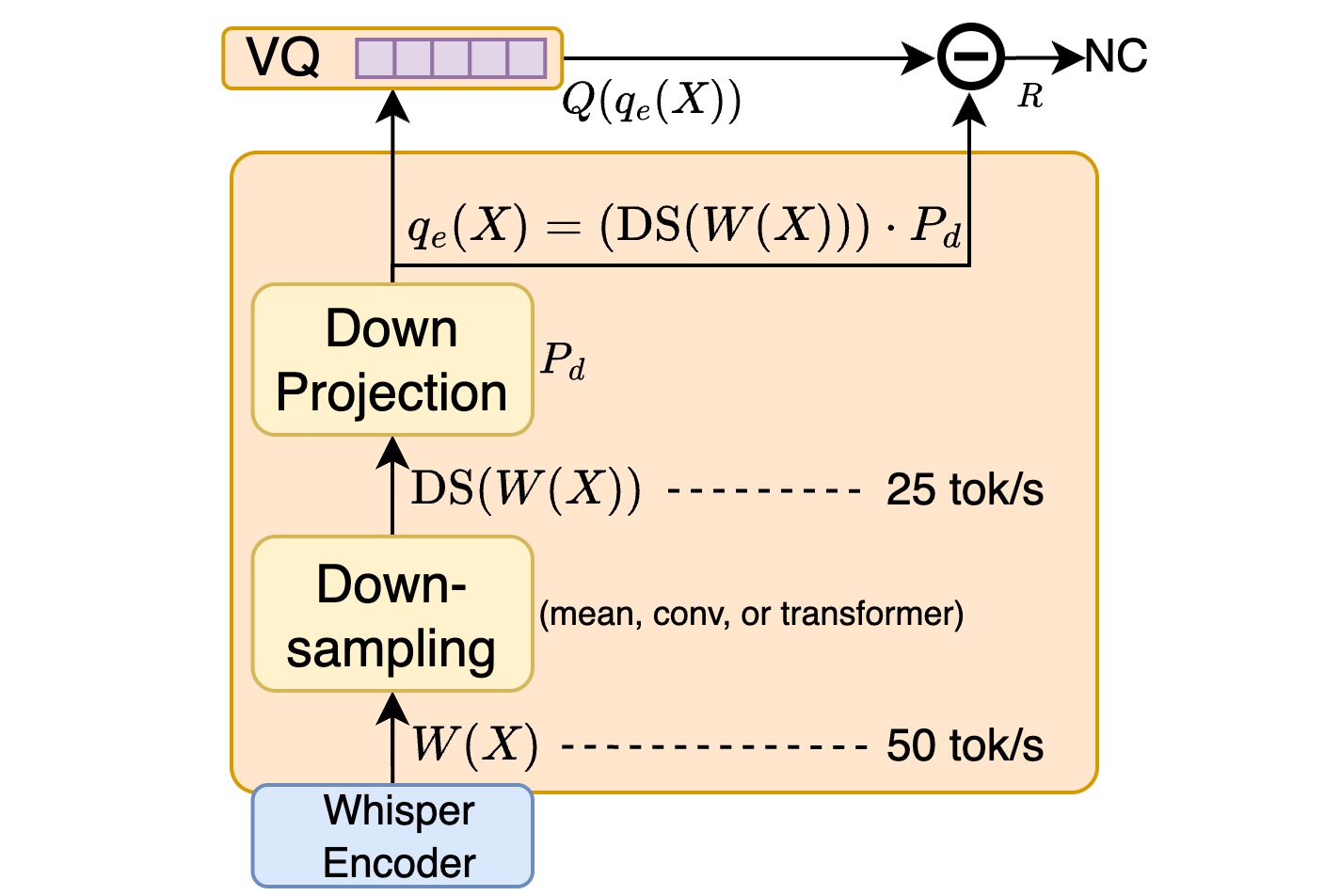}
    \caption{Latent encoder (orange). Whisper produces 50 embeddings/sec, while our encoder downsamples to 25 tokens/sec.}
    \label{arch_math}
    \vspace{-6 pt}
\end{figure}

As shown in Fig.~\ref{arch_math}, the latent encoder consists of a downsampling module, followed by a projection layer $P_d \in \mathbb{R}^{1024 \times 64}$. Downsampling reduces the temporal resolution by a factor of 2, such that each token encodes ~40ms of audio. In the english language, phonemes range from a duration of 80-120ms with an average around 100ms \cite{ma2022leveraging}. This implies that our tokens are sub-phonetic, and sequences of tokens make up phonemes and words. We experiment with mean-pooling, strided 1D convolution, and conv-transformer from \cite{zhang2022actionformer}, with the latter performing best. As trainable parameters in the encoder are important for embedding placement, for pooling and conv1d, we further use an MLP block. The output is seen in \eqref{le}.

\begin{equation}
    q_e(X) = \text{DS}(W(X)) \cdot P_d \label{le}
\end{equation}

\vspace{-10 pt}
\subsection{Vector Quantization}
\label{rvq}

VQ creates a discrete bottleneck that captures task-relevant semantic information, yielding quantized embeddings \( Q(q_e(X)) \in \mathbb{R}^{750 \times 64} \), derived from codebook \(\mathcal{C} \in \mathbb{R}^{N_{\text{CB}} \times 64} \) of size \( N_{\text{CB}}\). Here, \( q_e(X) \) denotes the output of the latent encoder. \( Q(\cdot) \) performs the nearest-neighbor lookup in \( \mathcal{C} \), whose codes represent denoised speech embeddings, which are further used for noise disentanglement and in the latent decoder. As a straight-through estimator is used, gradients will not directly impact codebook embeddings. Instead estimated moving averages are used to update the codebook with a codebook decay of 0.9. The optimal value of $N_{\text{CB}}$ varies with the size of the training corpus and its associated vocabulary. In our case $N_{\text{CB}} = 1024$ showed the best results.

Sequence information is preserved by 1) using fixed codebook token as padding token and 2) restoring positional encoding; This is not required when down-sampling using a conv-transformer as we find it is able to retain sequence information and embed all padded timesteps close together. 

\subsubsection{Noise Disentanglement}

Prior work shows that latent vector arithmetic can encode semantic differences—e.g., CSVAE \cite{klys2018learning} and attribute-aligned VAEs \cite{pati2021attribute}. We leverage this by treating the quantization residue $R$ as a representation of noise, and $Q(q_e(X))$ as capturing semantic content. This disentanglement is seen in \eqref{vq_subtraction}.

\begin{equation}
\setlength{\abovedisplayskip}{4pt}
\setlength{\belowdisplayskip}{4pt}
    R = q_e(X) - Q(q_e(X)) \label{vq_subtraction}
\end{equation}

We guide this separation using auxiliary loss terms that push $Q(q_e(X))$ toward clean semantics and $R$ toward noise. Inspired by \cite{gong2023whispernoise}, we apply a lightweight classifier (optionally transformer-augmented) to $R$ for noise classification. This supervision enforces extraction of noise features in $R$.

\subsection{Latent Decoder}

The latent decoder \( D_L(\cdot) \) reconstructs Whisper-like embeddings from the quantized outputs. It consists of a linear projection $P_u \in \mathbb{R}^{64 \times 1024}$, an up-sampling module, and a transformer refinement block. $P_u$ restores the embedding dimension, and the up-sampler restores sequence length $T'$. We compare repeat and transposed 1D convolution for up-sampling. The simpler repeat method, where tokens are duplicated in time, performed better in our setup (see Table~\ref{tab:vbdemand_eval_combined}). The transformer refinement block aids in smoothing artifacts due to up-sampling and aligning with Whisper's latent space.

\subsection{Training Cost Function}
\label{cost_func}

We employ a loss function that combines four components: ASR cross-entropy ($\mathcal{L}_{\text{ASR}}$), VQ commitment loss ($\mathcal{L}_\alpha$), a semantic disentanglement loss ($\mathcal{L}_\beta$), and an auxiliary noise classification loss ($\mathcal{L}_{\text{NC}}$). The total cost is defined as:

\vspace{-4pt}
\begin{equation}
\setlength{\abovedisplayskip}{4pt}
\setlength{\belowdisplayskip}{4pt}
    \mathcal{L} = \mathcal{W}_{\text{ASR}}\cdot\mathcal{L}_{\text{ASR}} + \mathcal{W}_\alpha\cdot\mathcal{L}_\alpha + \mathcal{W}_\beta\cdot\mathcal{L}_\beta + \mathcal{W}_{\text{NC}}\cdot\mathcal{L}_{\text{NC}} \label{total_loss}
\end{equation}

Through experiments, we found that $\mathcal{W}_\alpha + \mathcal{W}_\beta = 1.0$ gave the most consistent results, and setting $\mathcal{W}_{\text{ASR}} = 1.0$ provides a stable training direction early-on when quantizer outputs are still noisy. Table~\ref{tab:loss_ablation_fixed} shows the relevance of each loss term.

\subsubsection{ASR Cross-Entropy Loss}

Given a noisy speech signal $X$, which passes through the encoder, quantizer, and decoder to yield the model output $q_d(Q(q_e(X)))$, and a sequence of previous tokens $y_{t-1}, ..., y_1$, the cross-entropy loss aims to minimize the error in predicting the next token $y_t$:

\vspace{-4 pt}
\begin{equation}
\setlength{\abovedisplayskip}{4pt}
\setlength{\belowdisplayskip}{4pt}
    \mathcal{L}_{\text{ASR}} = -\sum_{t=1}^T \log \mathcal{P}_\theta(y_t | y_{t-1}, ..., y_1, q_d(Q(q_e(X)))) \label{asr_loss}
\end{equation}

This term is particularly important during the initial phases of training when quantizer embeddings are still adapting.

\subsubsection{Commitment Loss}

As introduced in foundational VQ-VAE works \cite{van2017vqvae}, the commitment loss prevents the uncontrolled growth of encoder outputs and stabilizes training. In our case, we define this loss as the L2 distance between the downsampled clean embeddings, $\text{DS}\left[q_e(X')\right]$ and the quantized outputs from the noisy signal:

\begin{equation}
\setlength{\abovedisplayskip}{4pt}
\setlength{\belowdisplayskip}{4pt}
    \mathcal{L}_\alpha = \left\|\text{DS}\left[q_e(X')\right] - Q(q_e(X)) \right\|_2^2 \label{alpha_loss}
\end{equation}

This encourages the noisy embeddings to stay close to their clean counterparts in latent space. The best performance was observed when $\mathcal{W}_\alpha = 0.5$.

\subsubsection{Semantic Disentanglement Loss}

This term corrects quantization artifacts by guiding the refined output toward the original Whisper latent space. It is defined as the L2 distance between the reference clean embeddings $q_e(X')$ and the output of the latent decoder:

\begin{equation}
\setlength{\abovedisplayskip}{4pt}
\setlength{\belowdisplayskip}{4pt}
    \mathcal{L}_\beta = \left\| q_e(X') - D_L(Q(q_e(X))) \right\|_2^2 \label{beta_loss}
\end{equation}

We found that $\mathcal{W}_\beta = 0.5$ worked best when paired with the commitment loss.

\subsubsection{Noise Classification Loss}

To explicitly model the noise component, we compute the quantization residue as seen in \eqref{vq_subtraction} and pass it through a lightweight classifier $NC(\cdot)$ to predict the noise class $Y'$. The loss is defined as:

\begin{equation}
\setlength{\abovedisplayskip}{4pt}
\setlength{\belowdisplayskip}{4pt}
    \mathcal{L}_{\text{NC}} = -\sum \log P_{\theta_{\text{NC}}}(Y' | NC(R)) \label{noise_loss}
\end{equation}

We found that scheduling $\mathcal{W}_{\text{NC}}$ inversely proportional to the classifier’s validation accuracy helped prevent overfitting and improved generalization.
\begin{table*}[t]
\caption{
ASR performance (WER) of baselines, our model, and upper bound. 
The first group varies codebook size with fixed mean-repeat (MR) sampling.
The second group compares various sampling configurations using the best codebook size ($N_\text{CB}=1024$.)}
\label{tab:vbdemand_eval_combined}
\centering
\renewcommand{\arraystretch}{1.1}
\setlength{\tabcolsep}{5pt} % Keep uniform spacing across all columns
\vspace{-8 pt}
\begin{tabular}{l|ccc|ccccc|cccccc|cc}
\toprule
\textbf{WER} $\downarrow$
& \multicolumn{3}{c|}{\textbf{Baseline}}
& \multicolumn{5}{c|}{\textbf{Codebook Size ($N_\text{CB}$) vs. WER}} 
& \multicolumn{6}{c|}{\textbf{Sampling Strategies vs. WER}} 
& \multicolumn{2}{c}{\textbf{Upper Bound}} \\
\cmidrule(lr){2-4} \cmidrule(lr){5-9} \cmidrule(lr){10-15} \cmidrule(lr){16-17}
\textbf{VBDemand} & Whisper & +Adapter & Speechless
& 128 & 256 & 512 & \textbf{1024} & 2048 
& MR & MC & CC & CR & \textbf{TR} & TC
& $\text{C}_\text{ZS}$ & $\text{C}_\text{TR}$ \\
\midrule
\textbf{Validation} & & & & & & & & & & & & & & & & \\
\textbf{all} & 6.75 & \textbf{0.35} & - 
                  & 0.58 & 0.31 & 0.29 & \textbf{0.21} & 0.49
                  & 0.21 &\textbf{0.19} & 0.24 & 0.28 & \textit{0.64} & 0.32 
                  & 2.47 & 0.68 \\
\midrule
\textbf{Test} & & & & & & & & & & & & & & & & \\
\textit{bus}        & 12.05 & 10.68 & -
                    & 2.58 & \textbf{1.86} & 1.99 & 2.67 & 3.24
                    & 2.67 & \textbf{1.62} & 2.91 & 2.10 & 1.86 & 1.86
                    & - & - \\
\textit{cafe}       & 25.71 & 4.03 & -
                    & 6.25 & 5.30 & 4.97 & \textbf{4.59} & 5.54
                    & 4.59 & \textbf{3.88} & 4.83 & 5.22 & \textbf{3.88} & 4.51
                    & - & - \\
\textit{living}     & 8.31 & 2.10 & - 
                    & 4.04 & 3.31 & \textbf{3.30} & 3.47 & 4.77 
                    & 3.47 & 4.03 & 3.07 & 3.63 & \textbf{2.75} & 3.72
                    & - & - \\
\textit{office}     & 9.01 & 1.18 & - 
                    & 2.51 & 2.04 & 1.93 & \textbf{1.57} & 2.98 
                    & \textbf{1.57} & 2.12 & 2.04 & 2.35 & 1.65 & 1.65
                    & - & - \\
\textit{pedestrian} & 13.44 & 1.15 & - 
                    & 3.30 & 2.15 & 2.14 & \textbf{1.84} & 2.30 
                    & 1.84 & 2.30 & 2.15 & \textbf{1.61} & 2.23 & 1.46 
                    & - & - \\
\textbf{all}  & 13.72 & \textbf{3.78} & 12.34
                    & 3.74 & 2.93 & 2.87 & \textbf{2.82} & 3.75 
                    & 2.82 & 2.79 & 2.99 & 2.98 & \textbf{2.47} & 2.63 
                    & 12.79 &  1.87 \\
\bottomrule
\end{tabular}

\vspace{0.6em}
\small\textit{Legend:}
\small{Whisper = Whisper-medium zero-shot, +Adapter = Whisper-medium with trainable linear adapter; }
MR = (mean, repeat), 
MC = (mean, conv1d), 
CC = (conv1d, conv1d), 
CR = (conv1d, repeat), 
TR = (conv-transformer, repeat), 
TC = (conv-transformer, conv1d);\\
$\text{C}_\text{ZS}$ = Whisper-medium zero-shot with \underline{C}lean speech input
$\text{C}_\text{TR}$ = Our model (1024-codebook TR setting)  with \underline{C}lean speech input
\vspace{-16 pt}
\end{table*}

\section{Experiments}

\subsection{Datasets}

\subsubsection{VBDemand}

VBDemand \cite{botinhao2016vbdemand} is widely used for speech enhancement and robust ASR \cite{hu2024large}. It includes clean-noisy speech pairs with transcriptions. We combine two subsets: one with 11,572 utterances (9.5h, English accents) and another with 21,225 utterances (19h, US and Scottish accents). Of this 28.5h, we reserve 1.5h (disjoint speakers) for validation and use the rest for training. Each training and validation sample contains 1 of 10 noise types. The noisy test set has 5 unseen noise types: bus, cafe, living room, office, and pedestrian.

\subsubsection{CHiME-4}
We use the \texttt{test-real} and \texttt{test-simu} splits from CHiME-4 \cite{vincent2016chime} for additional qualitative analysis to assess the generalization ability of our model under out-of-distribution noise and speaker conditions, as the clean-noisy pairs are perfectly time-aligned with various noise types.

\subsection{Setup}
All experiments were conducted on 2 A100 GPUs with a total batch size of 64 (32 per GPU). We used a linear learning rate warm-up for the first 500 steps, followed by a cosine annealing schedule. The initial learning rate was set to $1\text{e}^{-3}$, and optimization was performed using AdamW ($\beta_1 = 0.9$, $\beta_2 = 0.95$) until convergence. We initialize the whisper components using pretrained \textit{whisper-medium} checkpoints from the official OpenAI Whisper repository\footnote{\url{https://github.com/openai/whisper}}. Codebooks were initialized using Kaiming noise as described in \cite{dao2025speechless}. All WER results presented are derived using autoregressive decoding with a beam-size of 10 and temperature of 0.0.

\subsection{Evaluation}

\subsubsection{Metric}
As the Whisper decoder’s final objective is next-token prediction, we correlate ASR performance with the degree of semantic information retained in the latent embedding. We use word error rate (WER) as the primary evaluation metric for both our model and the baselines, and use relative error reduction (RER \%) to compare performances.

\subsubsection{Baseline}
We use the frozen \textit{whisper-medium} model on the VB-Demand test split to establish a performance baseline. Given that 1) Our method inserts adapter-style modules between the Whisper encoder and decoder without fine-tuning either and 2) a degree of semantic loss is inevitable in our approach due to quantization; We consider an additional baseline model (+Adapter in Table~\ref{tab:vbdemand_eval_combined}) where we insert trainable adapters—without discretization—between encoder and decoder. Each adapter block consists of 2 MLP layers with a GELU activation function between them. This black-box adapter provides a non-interpretable baseline approach that does not suffer info loss due to quantization. Lastly, the authors of Speechless \cite{dao2025speechless} show that VQ-VAE training with supervision only for semantic alignment results in poor performance in noisy conditions, hence we include results from their 2560-codebook model under similar conditions.

\subsubsection{Upper Bound}
We consider the performance with clean speech inputs as the upper-bound. The VBDemand clean test-set is used as the input and we tabulate zero-shot inference using Whisper-medium ($\text{C}_\text{ZS}$) as well as our best model's performance ($\text{C}_\text{TR}$: $N_\text{CB}=1024$ with conv-tranformer downsampling and repeat upsampling) with clean inputs.

\begin{table}[t]
\caption{Ablation study of loss terms. All configurations use Whisper-medium, $N_{\text{CB}} = 512$, and (mean, repeat) DS/US setup.}
\label{tab:loss_ablation_fixed}
\centering
\renewcommand{\arraystretch}{1.1}
\begin{tabular}{llccc}
\toprule
\textbf{Model} & \textbf{Losses Used} & \textbf{DS/US} & \textbf{$N_\text{CB}$} & \textbf{WER (\%)} $\downarrow$ \\
\midrule
\multirow{5}{*}{Ours}
& Only VQ and $\mathcal{L}_\text{ASR}$   & MR & 512 & 3.89 \\
& \hspace{1em}+ $\mathcal{L}_\text{NC}$      & MR & 512 & 3.63 \\
& \hspace{2em}+ Only $\mathcal{L}_\alpha$  & MR & 512 & 3.36 \\
& \hspace{2em}+ Only $\mathcal{L}_\beta$   & MR & 512 & 3.11 \\
& \hspace{2em}+ $\mathcal{L}_\alpha$ + $\mathcal{L}_\beta$ & MR & 512 & \textbf{2.87} \\
\bottomrule
\end{tabular}
\vspace{-4 pt}
\end{table}

\subsection{Primary Experiments}

We show that three factors contribute to the effectiveness of our method: (1) the VQ codebook size, (2) the number of trainable parameters before VQ, and (3) the choice of downsampling and upsampling strategies. (2) and (3) are tightly coupled as layers preceding the VQ module are expected to improve embedding placement and help the model learn meaningful representations while also performing downsampling.

In our first study, we fixed the downsampling to simple mean pooling and upsampling to repeat (MR), varying only the codebook size. After identifying $N_\text{CB} = 1024$ as optimal, we studied the effect of different sampling strategies on ASR and NC performance. These strategies are shown in Table~\ref{tab:vbdemand_eval_combined}.

Mean pooling relies heavily on MLP layers for embedding placement. Conv1D-based downsampling introduces additional learnable filters, while conv-transformer modules operate over both time and feature dimensions, enabling more complex token positioning. These results are presented in Table~\ref{tab:vbdemand_eval_combined}. We further discuss these results in detail in Section \ref{quant_obs}.
\vspace{-6pt}
\subsection{Ablation Study of Loss Terms}

To study the contribution of each loss term, we perform an experiment where loss components are added incrementally. This illustrates how each loss term contributes towards improving the model's performance. Results are shown in Table~\ref{tab:loss_ablation_fixed}.

%\newpage
\vspace{-4pt}
\section{Results and Discussion}

\begin{figure*}[t]
    \centering
    \includegraphics[width=0.85\linewidth]{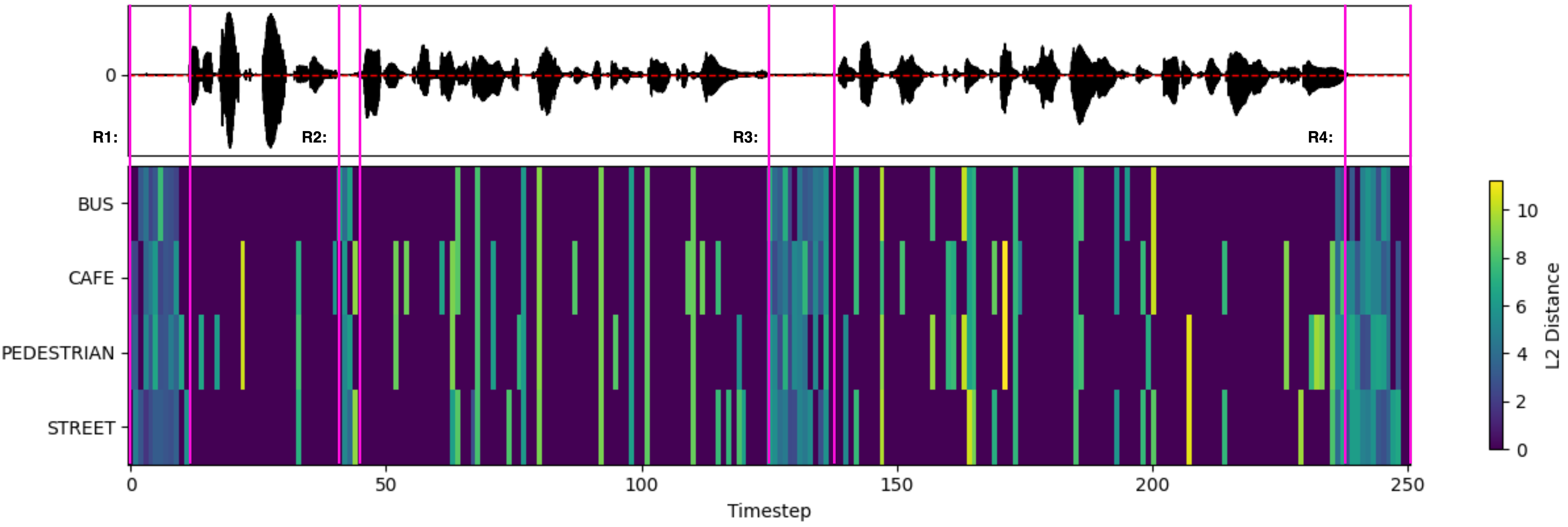}
    \vspace{-10 pt}
    \caption{Mean L2 Distance: Clean vs Noisy Embeddings: (top) Clean audio waveform from Chime-4 test dataset. The red dotted line represents an amplitude of 0. (bottom) The color at each position represents L2 distance of embeddings from our encoder between the clean signal and the clean signal mixed with various background noises. The y-axis indicates the mixed-in background noise, and the x-axis represents the time-steps. Lower values show that the encoder is able to generate more noise invariant representations for sub-word tokens and token sequences similar to those captured from clean speech.}
    \label{waveform}
    \vspace{-16 pt}
\end{figure*}

\begin{figure}[t]
    \centering
    \includegraphics[width=1.0\linewidth]{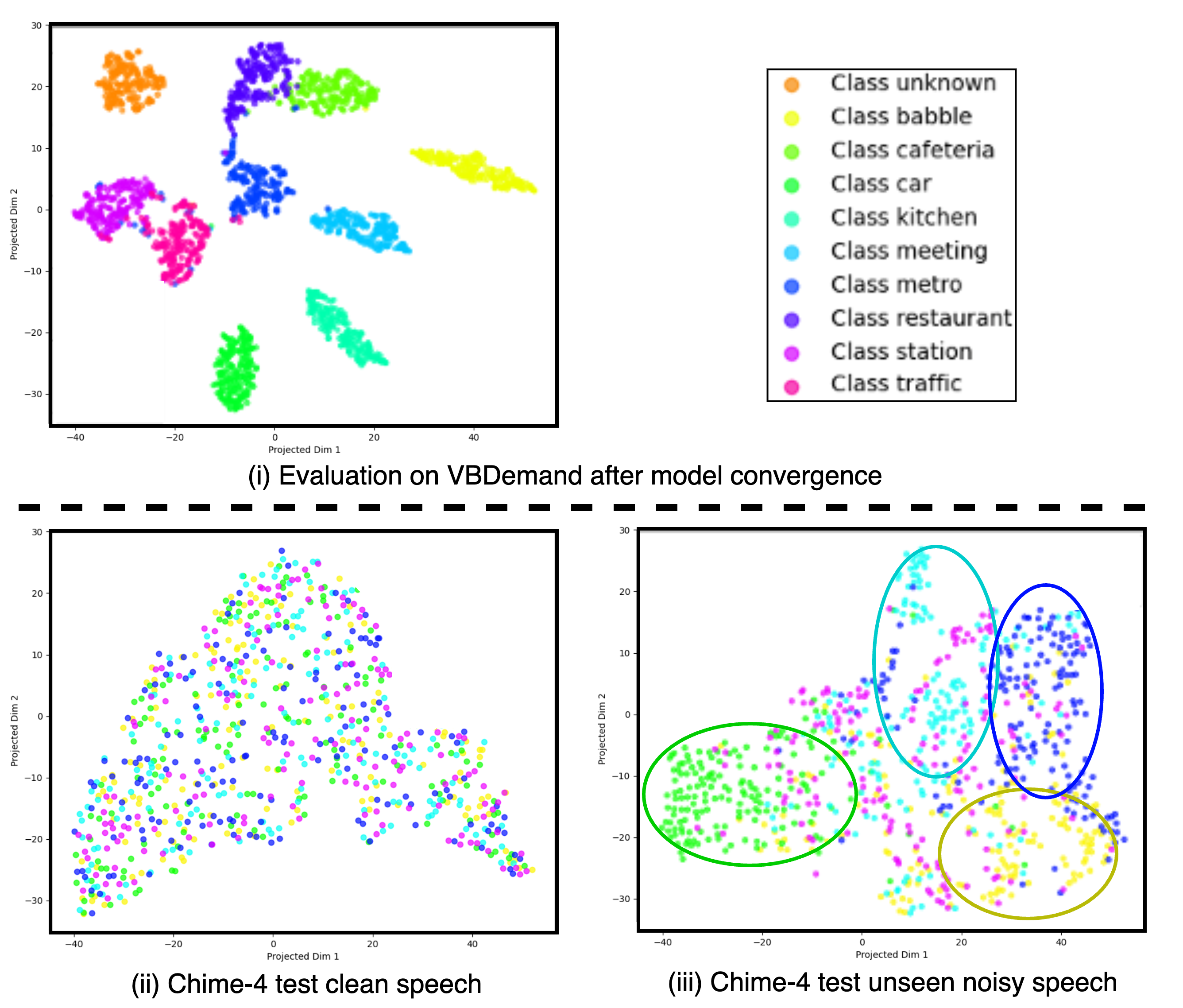}
    \vspace{-20 pt}
    \caption{
    (Top) T-SNE projections of penultimate-layer features from the noise classifier on VBDemand validation samples. Some samples were relabeled as "unknown" to encourage generalization. These noise types are seen during training. 
    (Bottom) Similar projections from clean and noisy CHiME-4 test samples. Clean speech yields uniform embeddings; noisy speech shows class-specific clustering. CHiME-4 noise types include café (green), bus (light blue), pedestrian (yellow), and street (dark blue), which are unseen during training.}
    \label{clusters}
    \vspace{-2 pt}
\end{figure}

\subsection{Quantitative Observations}
\label{quant_obs}

We present our findings in Table~\ref{tab:vbdemand_eval_combined}. The baseline whisper-medium model achieves a WER of 13.72\% on the VBDemand noisy test set. Adding trainable linear adapters further reduces the WER to 3.78\% (+Adapter; where no quantization info loss occurs). As shown, the majority of our configurations show performance gains compared to inserted linear MLP adapters, with our best approach ($N_\text{CB}$=1024, TR) achieving a WER of 2.47\%. This shows a relative error reduction of 82\% over the zero-shot whisper baseline, 35\% over the trainable adapter and 80\% over the base Speechless \cite{dao2025speechless} model. The conv-transformer method enables strong embedding placement requiring 28\% fewer epochs for convergence, while also showing the least amount of overfitting to the validation dataset.

Further, in Table~\ref{tab:loss_ablation_fixed}, we analyse the significance of each loss term. The first setting only employs ASR task supervision resulting in a WER of 3.89\% which is poorer than adapter training. An RER of 6.7\% is achieved by adding only noise disentanglement. The complete loss function results in a WER of 2.87\% and total RER of 20.9\%, showing that each term of our proposed loss function plays a significant role.

\subsection{Qualitative Observations} 

\subsubsection{Noise Interpretation}
Fig.~\ref{clusters} (top) shows T-SNE projections of representations from the penultimate layers of the noise-classifier of noisy samples from the validation set after model convergence. We observe that the information removed via the quantization error is clearly noise-class separable, and correlates directly with a distribution that helps the classifier categorize the background noise. The final noise classification accuracy for the VBDemand validation dataset is 98.23\%.

\vspace{-4pt}
\subsubsection{Clean vs. Noisy Speech}
Table \ref{tab:vbdemand_eval_combined} shows that our best model (TR) has a much lower WER of 1.87\% with clean speech inputs ($\text{C}_\text{TR}>\text{TR}$). This improved performance indicates that our model correctly encodes only semantic data, generalizing well to clean speech. Otherwise performance in clean speech would degrade due to information loss. We also perform inference on clean speech embeddings from the CHiME-4 dataset and as seen in Fig.~\ref{clusters} (bottom-ii), disentangled noise embeddings are not class-separable when the model is input with clean speech. Whereas, as seen in Fig.~\ref{clusters} (bottom-ii), even unseen noisy speech from CHiME-4 yields a good degree of class separability. These observations reinforce that our model separates out only valid noise information when available, while semantic information is correctly tokenized.

\subsubsection{Noise Invariance}
To examine the noise invariance of our latent representations, we examined embeddings of the same speech in the clean setting and with various noise samples mixed. Ideally the embeddings should be similar (invariant) at each time-step. To visualize this we make use of L2 distance of codebook embeddings of the clean and noisy speech. Fig.~\ref{waveform} shows both the clean audio sample from the CHiME-4 test dataset, along with L2 distances between each noisy sample and the original clean sample at each timestep. We see that the majority of embeddings and sequences of embedding show a high degree of noise invariance (dark purple). Patches of high embedding variance correlate directly with silence where the quantizer may not have enough information to generate useful embeddings (R1, R2, R3, R4). Localized errors occur across many noise types and seem to be due to fricative sounds and because CHiME-4 is outside of the training distribution.

\vspace{-2pt}
\section{Conclusions and Future Work}

In this work, we propose a novel disentanglement framework to improve the noise-robustness of discrete speech representations, by separating semantic speech tokens from background noise using vector quantization over Whisper embeddings. Without fine-tuning the Whisper encoder, our method achieves an 82\% error reduction compared to zero-shot Whisper and a 35\% improvement over adapter-based baselines, demonstrating the benefit of dual supervision over semantic and noise representations. Unlike prior methods such as Speechless that align only semantic content, our model jointly learns semantic and noise representations, yielding significantly better performance across noisy conditions (80\% reduction in WER). The quantization residue serves as an interpretable noise embedding, enabling accurate classification and an explainable latent space. Visual analysis confirms that the model generalizes well to unseen noise, while preserving clean speech performance. 

Future works can leverage the noise residue space for applications such as noise-guided training data augmentation, robust knowledge distillation, and enhancement of pre-trained discrete representation models.

% \begin{thebibliography}{1}

% \bibitem{1}
% G.~Eason, B.~Noble, and I.~N.~Sneddon, ``On certain integrals of
% Lipschitz-Hankel type involving products of Bessel functions,''
% \emph{Phil. Trans. Roy. Soc. London,} vol. A247, pp. 529-551, April
% 1955.

% \bibitem{2}
% J.~Clerk~Maxwell, \emph{A Treatise on Electricity and Magnetism,}
% 3$^{\rm rd}$ ed., vol. 2. Oxford: Clarendon, 1892, pp.68-73.

% \bibitem{3}
% I.~S.~Jacobs and C.~P.~Bean, ``Fine particles, thin films and exchange
% anisotropy,'' in \emph{Magnetism,} vol. III, G.T. Rado and H. Suhl,
% Eds. New York: Academic, 1963, pp. 271-350.

% \bibitem{4}
% K.~Elissa, ``Title of paper if known,'' unpublished.

% \bibitem{5}
% R.~Nicole, ``Title of paper with only first word capitalized,''
% \emph{J. Name Stand. Abbrev.,} in press.

% \bibitem{6}
% Y.~Yorozu, M.~Hirano, K.~Oka, and Y.~Tagawa, ``Electron spectroscopy
% studies on magneto-optical media and plastic substrate interface,''
% \emph{APSIPA Transl. J. Magn. Japan,} vol. 2, pp. 740-741, August 1987
% [\emph{Digests 9$^{\rm th}$ Annual Conf. Magnetics Japan,} p. 301,
% 1982].

% \bibitem{7}
% M.~Young, \emph{The Technical Writer's Handbook.} Mill Valley, CA:
% University Science, 1989.

% \end{thebibliography}

% \pagebreak

\section*{Acknowledgment}
The computational work for this article was partially performed on resources of the National Supercomputing Centre, Singapore (https://www.nscc.sg) and partially supported by the High Performance Computing Centre of Nanyang Technological University, Singapore.

\printbibliography

\end{document}